\documentclass[letterpaper, 10 pt, conference]{ieeeconf}  

\usepackage{hyperref}
\usepackage{amsmath,amssymb,amsfonts}
\usepackage{algorithmic}
\usepackage{multirow}
\usepackage{array}
\usepackage{graphicx}
\usepackage{textcomp}
\usepackage{xcolor}
\usepackage{booktabs}
\usepackage{bm}

\IEEEoverridecommandlockouts                              

\title{\LARGE \bf
Active propulsion noise shaping for multi-rotor aircraft localization
}

\author{Gabriele Serussi$^{1,*}$, Tamir Shor$^{1,*}$, Tom Hirshberg$^{1}$, Chaim Baskin$^{1}$, Alex M. Bronstein$^{1}$
\thanks{*Equal contribution}
\thanks{$^{1}$Technion -- Israel Institute of Technology, 3200003 Haifa, Israel
        {\tt\small \{gabrieles,tamir.shor,\}@campus.technion.ac.il \{chaimbaskin,bron\}@cs.technion.ac.il }}
\thanks{$^{2}$\url{https://github.com/tamirshor7/EARS_code}}
\thanks{$^{3}$\url{https://doi.org/10.7910/DVN/F0CVOQ}}}

\newcommand*\bb[1]{\bm{\mathrm{#1}}}

\begin{document}

\maketitle
\thispagestyle{empty}
\pagestyle{empty}

\begin{abstract}

Multi-rotor aerial autonomous vehicles (MAVs) primarily rely on vision for navigation purposes. However, visual localization and odometry techniques suffer from poor performance in low or direct sunlight, a limited field of view, and vulnerability to occlusions. Acoustic sensing can serve as a complementary or even alternative modality for vision in many situations, and it also has the added benefits of lower system cost and energy footprint, which is especially important for micro aircraft. This paper proposes actively controlling and shaping the aircraft propulsion noise generated by the rotors to benefit localization tasks, rather than considering it a harmful nuisance. We present a neural network architecture for self-noise-based localization in a known environment. We show that training it simultaneously with learning time-varying rotor phase modulation achieves accurate and robust localization. The proposed methods are evaluated using a computationally affordable simulation of MAV rotor noise in 2D acoustic environments that is fitted to real recordings of rotor pressure fields. Code$^{2}$ and data$^{3}$ are accompanied.

\end{abstract}

\section{Introduction}\label{sec:intro}

Research in the field of multi-rotor micro air vehicles (MAVs, colloquially known as ``drones") has been gaining increasing interest in recent years due to their rapidly growing applicability in a wide range of industries, such as agriculture, construction, and emergency services. This growth is enabled in part by the constantly improving ability of MAVs to operate autonomously in unknown and unexpected environments. A key element allowing this progress is the recent developments in artificial intelligence, enabling improved localization and navigation capabilities that are vital for the MAV to fulfill its designated tasks.  

Research in the field of MAV localization and navigation mainly focuses on employing various computer vision techniques to harness observed visual data into the MAV's decision-making process \cite{couturier2021review,khattar2021visual,krul2021visual,skoda2015camera,antonopoulos2022ros}. While these methods have proved to supply impressive performance, they are highly dependent on the availability and reliability of visual data. In cases of low visibility conditions, increased light exposure, occlusions, or visual-based adversarial attacks, visual localization may become ineffective. 

To overcome these difficulties, we turn to harnessing acoustic signals for MAV localization -- a domain that has been explored to a much lesser extent compared to its visual counterpart. In particular, we propose to focus on drone's self-noise generated by the propulsion system. Drones offer a limited amount of space for mounting sensors, and the demand for them to be autonomous requires minimizing their energy consumption as much as possible. The use of visual sensors, or even mounting speakers for the sake of sound generation, could be costly in this aspect. On the other hand, the drone's self-generated noise, which has so far been mainly considered a nuisance, is already generated for our disposal without any increased space consumption or costs. As we demonstrate in this study, the noise signal can be actively shaped to improve localization capabilities. This makes self-noise signals a viable candidate for acoustic-based localization. 

This paper makes the following contributions: 
Firstly, we introduce a novel neural network-based algorithm capable of localizing an MAV down to a few centimeters in a known acoustic environment using only the self-noise and the rotor angular positions as the inputs. 
Secondly, we propose a method for simultaneously optimizing the rotor phase modulation in concert with the localization model, obtaining a substantial improvement in localization accuracy. The learned phases are physically viable and do not interfere with the drone's flight stability. To the best of our knowledge, this is the first work to harness phase modulation for this purpose.
Lastly, we provide a fully-differentiable forward model of a drone in an acoustic environment and a first-of-its-kind set of recordings of a real rotor pressure field.

\section{Related Work}\label{sec:related_work}
Usage of acoustic signals in the field of robotics has proven effective in a variety of tasks and settings in recent years. \cite{fan2020acoustic} used auditory signals for joint localization and collision detection. Hu et al \cite{hu2011simultaneous} showed the potential of acoustic signals for the task of joint robot and sound source localization. Zhang et al\cite{zhang2021acousticfusion} aggregate acoustic signals from several dynamic sources to perform sound source localization. 

A number of works in particular have considered using auditory signals for the sake of localization alone. Eliakim et al \cite{eliakim2018fully} offered a sonar-based mechanism where a robot equipped with set of a speaker and a pair of mounted microphones learns to map the generated sound reflected into the microphones to location. Baxendale et al\cite{baxendale2018audio} harnessed Cerebellar models to perform audio based localization. Kim et al\cite{kim2012localization} localized in an underwater setting using an acoustic guided Particle Filter based algorithm. 

Several works have also used acoustic signals in multi-modal systems (\cite{vargas2021robust,franchi2019experimental}). These works consider acoustic signals alongside some other (mostly visual) signals from different channels, and integrate these channels to achieve the downstream target task.

The localization methods proposed in the above mentioned works are inherently dependant on some external set of speakers mounted on the drones or embedded into the environment. This dependence could be costly and limit the MAV's navigational flexibility. In our method we propose to replace these external signals with the sound emitted by the drone's rotors.

\section{Forward model}\label{sec:forward_model}
\begin{figure*}[bth]
    \centering
    \includegraphics[width=.9\linewidth]{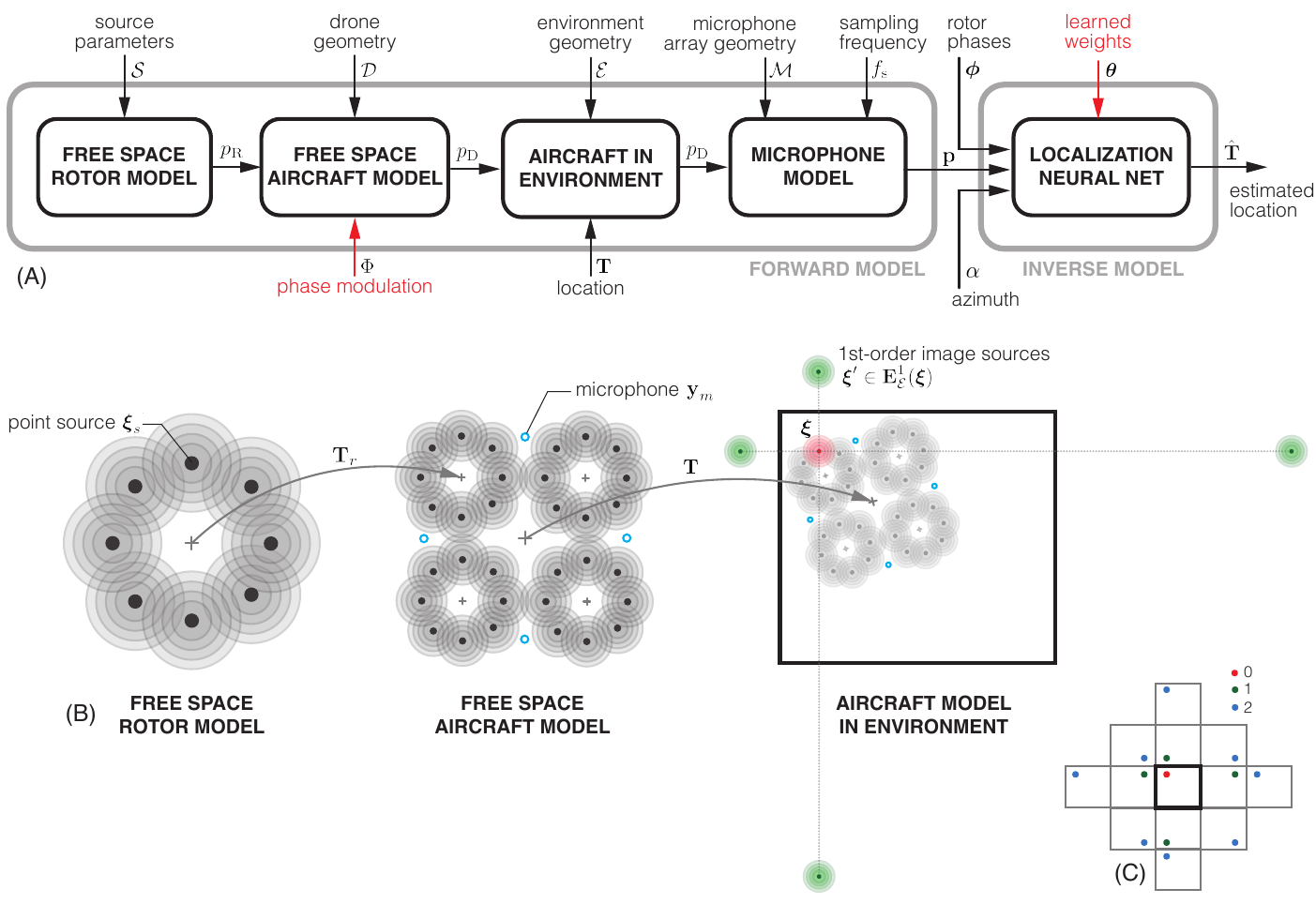}
    \caption{{\bf Forward and inverse models.} (A) Stages of the forward and the inverse models and their parameters. Learnable parameters are denoted in red. (B) The geometry of sources, microphones, and the environment. (C) The geometry of zeroth, first, and second-order image sources in a rectangular room.}
    \label{fig:model_diagram}
\end{figure*}
In what follows, we describe a fully-differentiable forward model of a multi-rotor aircraft in an acoustic environment. The need to model moving parts is avoided by using a phased array of fixed stationary sources; our experiments show that it allows us to accurately represent intricate pressure field geometries created by real MAV rotors. For a visualization of the model stages as well as for the definition of coordinate transformations, refer to Fig.~\ref{fig:model_diagram}.

\subsection{Rotor in free space} We model the pressure field generated by rotating rotor blades as a collection of fixed omnidirectional point sources located at a set of locations $\{\bb{\xi}_s\}$ (in rotor's coordinates) and temporally modulated with the signal $a_s(t)$ generated by source $s$ at time t:
\begin{equation}
a_s(t) = \sum_{k} \alpha_{sk} \cos(2k\omega t + \psi_{sk}), 
\label{eq:rotor_source}
\end{equation}
where $\omega$ is the shaft rotation frequency, $2$ corresponds to the modeled number of blades, the sum is over $K$ harmonics, and $\alpha_{sk}$ and $\psi_{sk}$ are, respectively, the amplitude and phase parameters of each harmonic $k$.
The pressure field generated by the point source at location $\bb{x}$ at time $t$ is given by the time convolution $a_s \ast h_0(\bullet, \bb{\xi}_s, \bb{x})$ with the free-space impulse response
\begin{equation}
h_0(t, \bb{\xi}_s, \bb{x}) = \frac{\delta( \omega t - \frac{1}{c} \| \bb{x} - \bb{\xi}_s \|  )}{4 \pi \| \bb{x} - \bb{\xi}_s \| },
\label{eq:rotor_pressure_fs}
\end{equation}
where $\delta$ is a Dirac delta, and $c$ denotes the speed of sound in air. 
The total rotor pressure field is given by
$$
p_\mathrm{R}(\bb{x}, t | \mathcal{S}) = \sum_{s} a_s(t) \ast h_0(t, \bb{\xi}_s, \bb{x}),
$$
where $\mathcal{S} = \{ \alpha_{sk}, \psi_{sk}, \bb{\xi}_s \}$ denote the model parameters. These parameters are fitted to a set of actual pressure measurements along concentric locations at different radii. Data collection and parameter fitting procedures are detailed in Section \ref{sec:dataset}. 

\subsection{Aircraft in free space} We model the pressure field of the entire drone rotor assembly of the aircraft by linear superposition of spatially-transformed and temporally-shifted pressure fields of the individual rotors. We denote by $\bb{T}_r$ the spatial transformation (rotation and translation) of the $r$-th rotor coordinates into aircraft coordinates, and by $\phi_r(t)$ the rotor's phase modulation. The total pressure field generated by the drone at location $\bb{x}$ (in aircraft coordinates) at time $t$ is given by
$$
p_\mathrm{D}(\bb{x}, t | \Phi, \mathcal{D}, \mathcal{S}) = \sum_r p_\mathrm{R}\left(\left.  \bb{x}, t - \frac{\phi_r (t)}{\omega} \right| \bb{T}_r \mathcal{S} \right),
$$
where we denote the phase modulations by $\Phi = \{ \phi_r \}$, the drone geometry parameters by $\mathcal{D} = \{\bb{T}_r \}$, and the transformed source parameters by $\bb{T} \mathcal{S} = \{ \alpha_{sk}, \psi_{sk}, \bb{T} \bb{\xi}_s \}$.

\subsection{Aircraft in acoustic environment} 
\label{subsec:env}
We model an acoustic environment by summing the contribution of the direct path (zeroth order) pressure field from the sources, their reflections from the walls (first order), the reflections of the reflections (second order), etc. Given a point source at location $\bb{\xi}$ (in environment coordinates), the environment geometry, denoted by $\mathcal{E}$, determines its map $\bb{E}^n_\mathcal{E}(\bb{\xi})$ to the set of $n$-th order image sources. 

Denoting by $\bb{T}$ the transformation of the aircraft coordinates to the environment coordinates, the drone pressure field at time $t$ and location $\bb{x}$ in the environment is given by 
$$
p_\mathrm{D}(\bb{x}, t | \bb{T}, \Phi, \mathcal{D}, \mathcal{S}, \mathcal{E}) = \sum_n \sum_{\mathcal{S}' \in \bb{E}^n_\mathcal{E} (\bb{T} \mathcal{S})}
p_\mathrm{D}(\bb{x}, t | \Phi, \mathcal{D}, \mathcal{S}'),
$$
where $\bb{E}^n_\mathcal{E} (\mathcal{S}) = \{ \gamma^n \alpha_{sk}, \psi_{sk}, \bb{E}^n_\mathcal{E} (\bb{\xi}_s) \}$, and $\gamma$ is the acoustic reflection coefficient according to which higher-order decay exponentially due to acoustic energy absorption in the wall material. 

\subsection{Microphone array} Denoting by $\mathcal{M} = \{\bb{y}_m\}$ the locations of $M$ omni-directional microphones (in aircraft coordinates), the measurement of the $m$-th microphone of the pressure field created by the drone in the environment at time $t$ is given by
$$
p_m(t| \bb{T}, \Phi, \mathcal{D}, \mathcal{M}, \mathcal{S}, \mathcal{E}) = p_\mathrm{D}(\bb{T} \bb{y}_m, t | \bb{T}, \Phi, \mathcal{D}, \mathcal{S}, \mathcal{E}) \ast h_\mathrm{AA} (t),
$$
where $h_\mathrm{AA}$ is the impulse response of the anti-aliasing low pass filter matching the microphone's sampling frequency $f_\mathrm{s}$. We collectively denote all microphone readings in discrete time by $\bb{p}[n] = (p_1(n/f_\mathrm{s}),\dots, p_M(n/f_\mathrm{s}) )$.

Our JAX-based implementation of the forward model based on the {\tt pyroomacoustics} package allows to differentiate its output with respect to the parameters. In Section \ref{sec:learning_phase_modulation}, we specifically use the gradients with respect to the rotor phases $\Phi$ to learn optimal phase modulations.

\section{Inverse model}\label{sec:inverse_model}
The localization inverse problem consists of estimating the spatial orientation and location $\bb{T}=(\bb{R},\bb{t})$ from the microphone readings $\bb{p}$, assuming known the forward model. Since the rotor phases are controlled on a best-effort basis by a flight controller that also needs to ensure a stable flight in the presence of perturbations such as wind, we also assume the phases are measured continuously and provided as the input sampled by the rotor encoders with the frequency $f_e$, yielding $\bb{\phi}[n] = (\phi_1(n/f_\mathrm{e}),\dots, \phi_4(n/f_\mathrm{e}) )$.

In this study, we restrict our attention to the estimation of the location parameter $\bb{t}$ only, assuming the orientation $\bb{R}$ is known and provided externally (e.g., from a compass sensor). We also defer to future studies the more challenging setting of simultaneous localization and mapping, in which the environment $\mathcal{E}$ needs to be estimated together with $\bb{t}$. Under these assumptions, we denote the inverse operator as
 $\hat{\bb{t}}(\bb{p}, \bb{\phi} | \alpha)$, representing the orientation as the azimuth $\alpha$ and omitting for clarity the dependence on the source, drone, and environment geometries that are assumed fixed and known.
 
\subsection{Localization model}\label{subsec:localization_model}
We model the inverse operator as a feed-forward neural network receiving the sampled microphone recordings $\bb{p}$ and the azimuth $\alpha$, and outputting a vector of location parameters. We used two separate trainable positional embeddings: one for the time dimension allowing the model to distinguish the data at different time locations, and another encoding the microphone that perceived the relevant input sound sample. This allows the model to recognize the source of the pressure field.  Microphone readings are transformed to the short-time Fourier transform (STFT) domain and represented as magnitude and phase. These embeddings are summed to the STFT frames which are encoded by a 3D convolutional layer followed by a Transformer-Encoder architecture \cite{vaswani2023attention}. The azimuth $\alpha$ is represented by its sine and its cosine, and these latter are encoded by an MLP. The encoded $\bb{p}$ and $\alpha$ are first concatenated and then aggregated using an MLP followed by a Transformer-Encoder architecture which returns an estimate of the location. The knowledge of the forward model is implicit through training detailed in the sequel.

\begin{figure*}[tb]
        \vspace{-0.6cm}
        \centering
        \includegraphics[width=\linewidth]{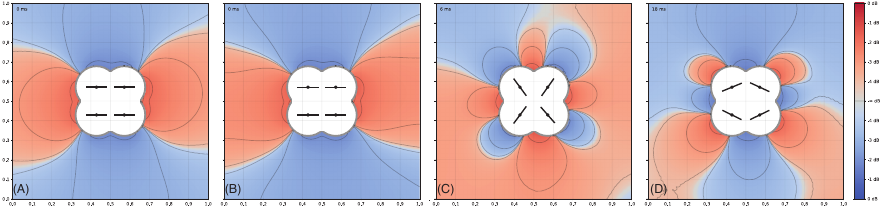}
        \caption{{\bf Simulated pressure fields} generated by the aircraft in free space (A) and in a square room at different times (B-D). Positive and negative pressures are color-coded in red and blue, respectively. A circle of $0.51$m around each rotor is not modeled in the absence of data recording in blade proximity.}
    \label{fig:fields}
\end{figure*}
\subsection{Model training}\label{subsec:model_training}
The model is trained by minimizing the loss
\begin{equation}
\mathbb{E}_{\bb{t}, \alpha} \,  \left\|
\bb{t} - \hat{\bb{t}}_{\bb{\theta}}( \bb{p}((\bb{R}_\alpha,\bb{t}), \Phi), \bb{\phi}(\Phi) | \alpha ) \right\|^2,
\label{eq:loss}
\end{equation}
where $\|\bb{t}- \hat{\bb{t}}\|^2$ quantifies the localization error, $\bb{p}((\bb{R}_\alpha,\bb{t}), \bb{\phi})$ denotes the forward operator simulating the microphone readings given the aircraft location and orientation $(\bb{R}_\alpha,\bb{t})$ and the rotor phases $\Phi$, and $\bb{\phi}(\Phi)$ denotes the sampling of the phases. For notation clarity, we omit the dependence on the known geometries. 
The expectation is approximated on a training set of random viable aircraft locations and orientations in the environment. Optimization is performed over the localization model parameters collectively denoted as $\bb{\theta}$.

\section{Learning rotor phase modulation}
\label{sec:learning_phase_modulation}

Among the ``hardware'' properties of the forward model (like the drone geometry), the rotor phase modulation, $\Phi$, is freely controllable, at least in principle. Differences in relative rotor phases exhibit a dramatic impact on the pressure field generated by the aircraft while being inconsequential to its flight characteristics. Changing the acoustic field generated by the drone at a static location is essentially synonymous with performing measurements through distinct forward models, potentially providing more information useful for localization. These facts make the phase modulation an appetible degree of freedom to try optimizing simultaneously with the inverse model training. The corresponding minimization of (\ref{eq:loss}) can be extended as
\begin{equation}
\min_{\bb{\theta}, \Phi} \mathbb{E}_{\bb{t},\alpha} \left\|
\bb{t} - \hat{\bb{t}}_{\bb{\theta}}( \bb{p}((\bb{R}_\alpha,\bb{t}), \Phi), \bb{\phi}(\Phi) | \alpha ) \right\|^2 + \ell_\mathrm{phys}(\Phi),
\label{eq:phase_loss}
\end{equation}
(note $\Phi$ among the optimization variables), with the additional second term $\ell_\mathrm{phys}(\Phi)$ that imposes physical constraints on the learned phases. In what follows, we describe the details of this learning problem.
\subsection{Parametrization}\label{subsec:parametrization}

The solution of (\ref{eq:phase_loss}) requires representing the continuous rotor phase modulation functions, $\phi_r(t)$, as a finite set of discrete parameters amenable to optimization.
The angular position of a rotor at time $t$ is given by $\omega t + \phi(t)$, suggesting that $\omega + \dot{\phi}(t)$ determines the instantaneous angular velocity. We therefore opted for representing the temporal derivative directly and obtaining the phase $\phi(t)$ through integration. We further assume that the phase modulation signal is periodic with some period $T_\mathrm{p}$ which, for convenience, we set to be an integer multiple of nominal revolution periods $2 \pi / \omega$ ($T_\mathrm{p} = 16 \pi / \omega$ in our experiments). 
We parametrize the phase derivative in the basis of $K$ discrete cosine harmonics,
\begin{equation}
\dot{\phi}(t) = \sum_{k > 0} \beta_k \cos\left( \frac{2\pi k t}{T_\mathrm{p}} \right), 
\label{eq:dphi}
\end{equation}
such that 
\begin{equation}
\phi(t) = \sum_{k > 0} \frac{\beta_k}{k} \sin\left( \frac{2\pi k t}{T_\mathrm{p}} \right). 
\label{eq:phi}
\end{equation}
With some abuse of notation, we continue to collectively denote by $\Phi = \{ \beta_{rk} \}$ the parameters characterizing the phase modulations of all rotors. 
\subsection{Physical constraints}\label{subsec:physical_constraint}
In order to guarantee that the found phase modulations are actually realizable on a real aircraft, every rotor's phase has to be subjected to a set of physical constraints that are implemented as penalty terms in the training loss (\ref{eq:phase_loss}).

\paragraph*{Angular velocity constraint} keeps the instantaneous angular velocity within the range $[-\omega_\mathrm{max}, \omega_\mathrm{max}]$. This is achieved by imposing a hinge penalty in the form
\begin{equation}
\ell_\omega = \sum_{t} [\dot{\phi}(t) + \omega - \omega_\mathrm{max}]_+ + [-\omega_\mathrm{max} - \dot{\phi}(t) - \omega ]_+,
\label{eq:penalty-omega}
\end{equation}
where $[\omega]_+ = \max\{\omega, 0\}$ and the sum is over a discrete set of times in the interval $[0, T_\mathrm{p}]$. The phase derivative is directly accessible in closed form according to (\ref{eq:dphi}).

\paragraph*{Angular acceleration constraint} keeps the instantaneous angular acceleration within the range $[-\alpha_\mathrm{max}, \alpha_\mathrm{max}]$. As before, the constraint is translated into the penalty
\begin{equation}
\ell_\alpha = \sum_{t} [\ddot{\phi}(t) - \alpha_\mathrm{max}]_+ + [-\alpha_\mathrm{max} - \ddot{\phi}(t)  ]_+,
\label{eq:penalty-acc}
\end{equation}
where the phase second-order derivative is also given in closed form,
\begin{equation}
\ddot{\phi}(t) = \sum_{k \ge 0} k \beta_k \sin\left( \frac{2\pi k t}{T_\mathrm{p}} \right).
\label{eq:ddphi}
\end{equation}

\paragraph*{Zero net thrust constraint} Since the rotor's angular velocity is linearly related to the amount of thrust it produces, in order not to interfere with aircraft stability, we demand that the net change in $\dot{\phi}(t)$ over a sufficiently long period of time is zero. Since the phases are represented directly as harmonic series, it is convenient to impose zero net thrust constraints by penalizing the energy contained in the low frequencies of the phase. This is achieved through a penalty of the form
\begin{equation}
\ell_\mathrm{thrust} = \sum_{k > 0} G(k) \beta_k^2,
\end{equation}
where $G(k)$ is a low-pass kernel monotonically decreasing with frequency. In our experiments, we used a sum of Gaussian kernels with varying bandwidth. Note that by construction, $\dot{\phi}(t)$ integrates to zero over the entire period $[0,T_\mathrm{p}]$.

The aforementioned physical constraints are further summed over all rotors and combined into a single penalty term with relative weights of $\lambda_{\omega}=0.1$, $\lambda_{\alpha}=0.1$, $\lambda_\mathrm{thrust} = 1$ set to the angular velocity, acceleration, and zero net thrust terms, respectively.

\begin{figure*}[tbh]
        \centering
        \includegraphics[width=0.95\linewidth]{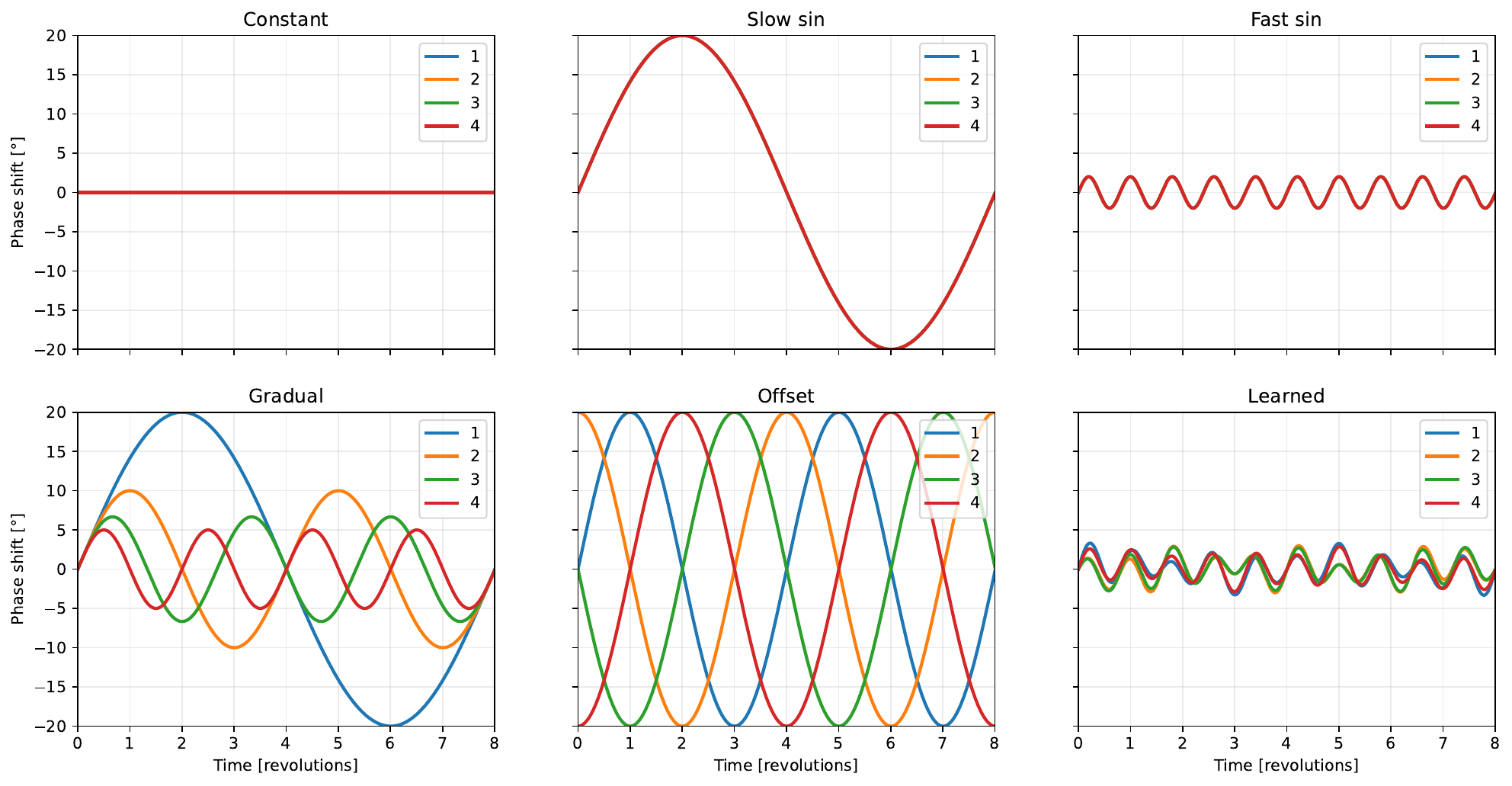}
        \caption{{\bf Rotor phase modulations} evaluated in the experiments. Rotors are color-coded. Counter-rotating rotor pairs are (1,4) and (2,3).}
    \label{fig:phases}
\end{figure*}

\subsection{Phase modulation optimization}\label{subsec:phase_modulation_optimization}

Utilizing the differentiability of both the forward and inverse models, the loss (\ref{eq:loss}) is backpropagated through both models to jointly update the localization model parameters $\bb{\theta}$ as well as the phase parameters $\Phi = \{ \beta_{rk}\}$. The localization model is extended by taking as input also $\Phi$ which is embedded using two trainable positional embeddings: one for the time dimension, and another encoding the rotor $r$ whose phase is modulated. Similarly to the sampled microphone recordings $\bb{p}$, the phase modulations are transformed to the STFT domain and represented as magnitude and phase. The embeddings are summed to the STFT frames which are encoded using a 3D convolutional layer followed by a Transformer-Encoder architecture. Downstream of the Transformer-Encoder these encodings are concatenated to the encodings of $\bb{p}$ and $\alpha$, which are fed to an MLP followed by a Transformer-Encoder which outputs the location prediction $\bb{t}$.

To improve convergence, we adopted the ``freezing`` technique similar to the one we previously used in \cite{shor2023multi} for the simultaneous learning of scan trajectories and reconstruction operators in magnetic resonance imaging. According to this method, each of the rotor phases are learned separately for several epochs, while keeping ``frozen'' the phases of the rest of the rotors. This is followed by jointly fine-tuning all rotor phases at once for a certain number of epochs. During the process, the localization model parameters $\bb{\theta}$ are always updated.

\section{Multi-measurement aggregation}

Due to environment symmetries, the inverse operator $\hat{\bb{t}}(\bb{p}, \bb{\phi} | \alpha)$ tends to have high uncertainties for a specific set of orientations. 
To mitigate it, we collect and aggregate multiple measurements from different orientations.
Let us assume that $J$ measurements are acquired at the same latent location $\bb{t}$ at a known set of orientations $\alpha_1,\dots,\alpha_J$, resulting in matrices of microphone and rotor phase readings, $\bb{P} = (\bb{p}_1,\dots,\bb{p}_J)$ and $\bb{\Phi}=(\bb{\phi}_1,\dots\bb{\phi}_J)$, with $\bb{p}_j = \bb{p}((\bb{R}_{\alpha_j},\bb{t}), \bb{\phi}_j)$. 
We then estimate the location parameter $\hat{t}_j = \hat{\bb{t}}(\bb{p}_j, \bb{\phi}_j | \alpha_j)$ separately from each measurement and aggregate the estimates by calculating their geometric median
\begin{equation}
\hat{\bb{t}}  = \mathrm{arg}\min_{\bb{t}} \sum_j \| \bb{t} - \hat{\bb{t}}_j \|.
\label{eq:median}
\end{equation}
The latter is calculated using the Weiszfeld's algorithm \cite{weiszfeld1937geometricmedian}, typically taking a few iterations to converge.


\begin{figure*}[tbh]
        \centering
        \includegraphics[width=0.95\linewidth]{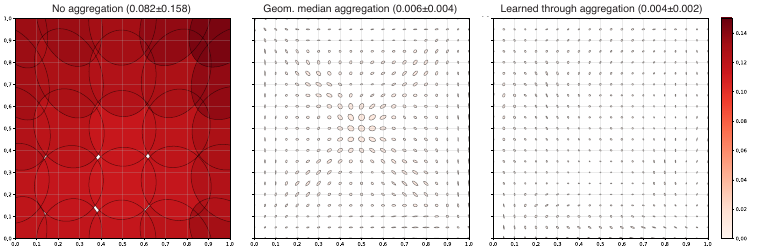}
        \caption{{\bf Localization uncertainty} in a square $5$m $\times$ $5$m room with learned phase modulation. Shown are $1 \sigma$ uncertainty ellipses calculated in a $0.05$ radius over a uniform grid of $64$ azimuthal orientations. RMS errors are color-coded. Left-to-right: no angular aggregation; geometric median aggregation post-training; and training through the aggregation. Average RMS localization accuracy is reported in the captions in relative units. }
    \label{fig:loc_uncertainty}
\end{figure*}

\begin{figure*}[tb]
        \centering
        \includegraphics[width=1\linewidth]{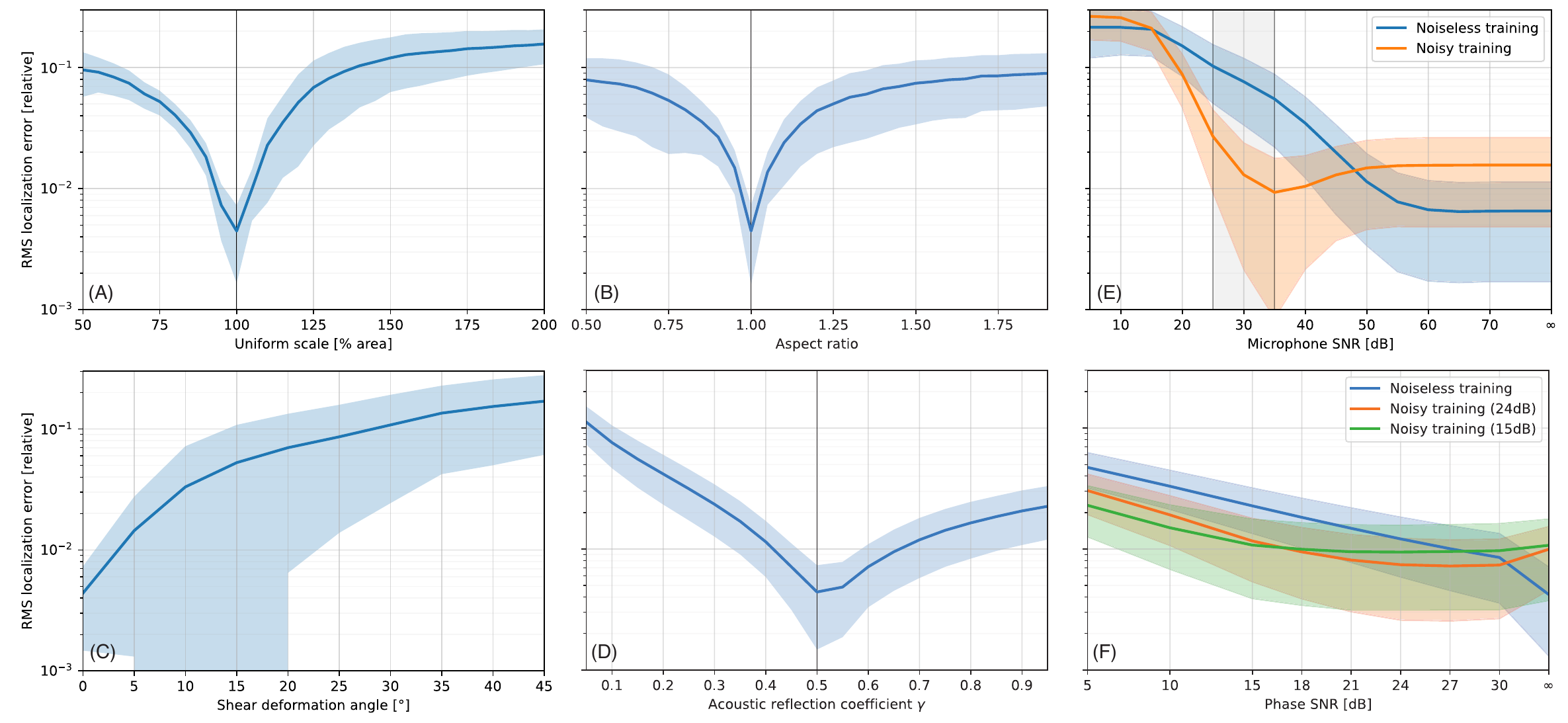}
        \caption{{\bf Robustness to various sources of modeling and sensing noise}. (A-D) environment parameters mismatch between training and inference. Nominal parameters are indicated by vertical lines.  (E-F) sensitivity to sensing and rotor phase noise. Shown is the performance of a model trained in noiseless settings compared to a model trained with noise injection. Shaded regions indicate $1 \sigma$ confidence intervals calculated over a uniform grid of locations in the room. A $5$m $\times$ $5$m room was used at training. Phase modulation was trained in all models.   }
    \label{fig:robustness}
\end{figure*}

\section{Experimental evaluation}\label{sec:experimental_evaluation}

In what follows, we present a simulation evaluation of the performance of the proposed methods, with real free-space recordings of an MAV rotor.

\subsection{Single rotor data acquisition}\label{sec:dataset}
Existing publicly available audio datasets of MAV and single rotors are few, and mainly consist of flyover scenarios only, making the recordings vulnerable to aircraft movements and external environmental disturbances, such as wind \cite{strauss2018dregon}. Therefore, to model the self sound of a rotor in free-space, we recorded a new dataset of a single spinning rotor in a semi-anechoic room. The recording setup included a motor with a rotor mounted on a tripod placed in the middle of the room. A microphone array of four RODE NTG4 directional shotgun microphones was placed circularly around the rotor to capture the sound. To measure the instantaneous shaft position, an encoder was mounted on the motor, and its readings were synchronized with the array recordings using the Roland OCTA-CAPTURE digitizer at a $44.1$kHZ sampling rate for the audio, and $128$ samples per revolution for the encoder. The four microphones were placed with $90$ degree angular steps from each other at eight radial locations from the rotor axis: $0.53$, $0.57$, $0.63$, $0.68$, $0.73$, $0.83$, $0.93$, and $1.03$ meters.
An open-loop control system was used to control the motor speed. The control hardware included a BeagleBoard with an Electronic Speed Controller (ESC) providing up to $40$ amperes of current to the motor. In each
experiment, the motor was stabilized at $10$ fixed angular velocities for the duration of $5$ seconds. The angular velocity was measured through the encoder readings. 

\subsection{Simulation settings}\label{sec:exp_simstudy}

The rotor source was modeled according to (\ref{eq:rotor_pressure_fs}) with $256$ point sources with locations $\bb{\xi}_\mathrm{s}$ arranged into two concentric circles at radii $0.23$m and $0.51$m, each containing $128$ points spread at a uniform angular grid. Each point source was modeled according to (\ref{eq:rotor_source}) with four harmonics $k=0.5, 1, 2, 3$ harmonics (the ``half" harmonic was used to capture the mechanical noise produced by the motor itself). The total of $2048$ parameters were fitted to the recorded data by solving a non-linear least-squares problem using L-BFGS.  

We used the two-dimensional forward model detailed in Section~\ref{sec:forward_model} to simulate the pressure fields created by a four-rotor aircraft in a rectangular room. Unless specified otherwise, all experiments were performed in a $5$m $\times$ $5$m room with wall acoustic reflection coefficient $\gamma = 0.5$. In this room we considered only the positions that could be physically occupied by the drone, namely, we took a margin of $0.93$ m from each wall. Reflections were calculated according to Section~\ref{subsec:env} up to the first order. The rotors were placed in a square formation $1.42$m apart, with the forward left and rear right rotors rotating clockwise, while the forward right and the rear left rotors rotating counter-clockwise. The baseline angular velocity was set to $\omega = 23.46$ rotations per second (RPS). The sensing array comprised $8$ microphones circularly arranged at a radius of $0.91$m from the drone center with an equal angular spacing of $45$ degrees.  

\subsection{Training settings} \label{sec:exp_training}

Training and evaluation were performed on a single NVIDIA GeForce RTX 2080 GPU. Optimization in all experiments was done using the Adam optimizer \cite{kingma2014adam}. For the localization model, we used a 3D convolutional layer with a kernel size and a stride of (3,3,2), a 3-layer Transformer encoder with a $1024$ hidden dimension and a single output head. The Transformer encoder's weights were trained with the learning rate of $10^{-5}$.
To learn the phase modulation, we used a basis of $K=10$ discrete cosine harmonics in (\ref{eq:phi}). Phase coefficients $\beta_{rk}$ were learned individually for each rotor using Adam, with an initial learning rate of $0.001$ and decay rate of $0.5$ every $20$ epochs, starting from the optimization fine-tuning stage. $160$ epochs were used in all training runs with a batch size of $50$. These 160 epochs were split into four $25$-epoch cycles of per-rotor phase optimization followed by $25$ epochs of joint optimization for all modulations. Finally, phase parameters were frozen and the localization model was trained for $35$ additional epochs.

The following physical constraints were imposed as described in Section~\ref{subsec:physical_constraint}: $\omega_\mathrm{max}=8000$ rad/sec for the angular velocity constraint (\ref{eq:penalty-omega}); $\alpha_\mathrm{max}=4000$ rad/sec$^2$ for the angular acceleration constraint (\ref{eq:penalty-acc}). 
Each room has been sampled at $3969$ spatial points with $64$ orientations. This dataset was split into train, validation, and test sets by the ratios of $80\%$, $10\%$, and $10\%$, respectively. For each location and orientation, an input of $1025$ time steps spanning eight rotor revolutions (about $0.34$ sec) was generated.

\subsection{Impact of rotor phase modulation}\label{subsec:comparative_experiments}

\begin{figure}[t]
        \centering
        \includegraphics[width=\linewidth]{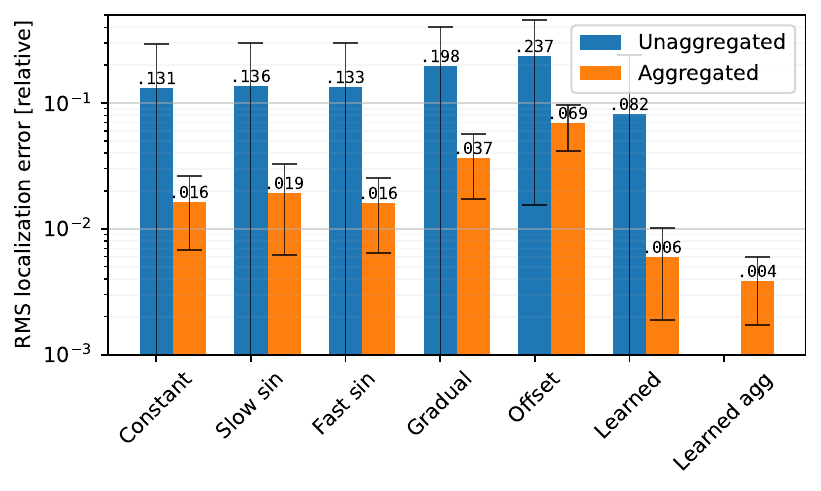}
        \vspace{-7mm}
        \caption{{\bf Localization accuracy of different phase modulations} in a $5$m $\times$ $5$m room. RMS errors are reported in relative units. $1\sigma$ confidence intervals were calculated over all room locations (and orientations in the unaggregated case). }
    \label{fig:loc_phases}
\end{figure}

This set of experiments is designed to evaluate the extent to which phase modulation learning helps achieve superior localization accuracy. To this end, we compared our learned per-rotor phase modulations with a set of constant modulations, where the pairwise phase differences between the rotors are fixed in time, and with a set of handcrafted modulations where the phase differences vary in time. All learned, handcrafted, and constant modulations fully satisfied the physical constraints.
The following modulations were evaluated (refer to Fig~\ref{fig:phases}):
\begin{enumerate}

\item \emph{Constant} -- all rotors at constant phase $0$.

\item \emph{Slow sine} -- for all rotors a sine wave with a period of $8$ rotor revolutions and peak amplitude of $20$ degrees.

\item \emph{Fast sine} -- for all rotors a sine wave for all rotors completing $10$ periods per $8$ revolutions and peak amplitude of $2$ degrees.

\item \emph{Gradual freq.} -- a sine wave with a period of $8$ rotor revolutions and peak amplitude of $20$ degrees for the first rotor. For each rotor $r$, the frequency of the sine is increased by $r$ and the amplitude is decreased by $r$. 

\item \emph{Offset}  -- sine waves of $8$ rotor revolutions and peak amplitude of $20$ degrees, offset by multiples of $90$ degrees for each rotor.

\item \emph{Learned} -- phases learned as described in Section~\ref{sec:learning_phase_modulation}.

\end{enumerate}
Localization accuracy of the different phase modulations is summarized in Figure~\ref{fig:loc_phases}. Phase learning improves localization by over a factor of $\times 2.7$ compared to the best hand-crafted phase. Figure \ref{fig:loc_uncertainty} visualizes the spatial distribution of the localization error using the learned phases with and without angular aggregation using the geometric median (\ref{eq:median}). We also compare phase modulation learned through the aggregation step. In all cases, $64$ orientations were aggregated. Our conclusion is that aggregation has a dramatic (over $\times 13$) effect on localization accuracy. Learning through the localization model brings an additional $\times 1.5$ improvement, further characterized by a spatially more uniform error distribution. 

\subsection{Robustness to environment modeling errors} 
We conducted several tests to assess the sensitivity of the model to the presence of different sources of environment modeling errors. To that end, the localization model and the rotor phases were trained on a nominal environment, while a perturbed environment was presented at evaluation time. The following parameters were perturbed in isolation: 
\begin{enumerate}

\item \emph{Uniform room scaling} by factors ranging from $0.5$ to $2$ of area (nominal: $1$).

\item \emph{Room aspect ratio} ranging from $0.5$ to $2$ while preserving the room area (nominal: $1$).

\item \emph{Room shear deformation} transforming the square room into a parallelogram by changing its right angle with a deformation ranging from $0$ (nominal) to $45$ degrees (maximum deformation).

\item \emph{Acoustic reflection coefficient $\gamma$} ranging from $0.05$ to $0.95$ (nominal: $\gamma = 0.5$).
\end{enumerate}
Localization accuracy in response to these perturbations is depicted in Figure \ref{fig:robustness} (A-D). In general, we can conclude that the model can gracefully cope with over $20\%$ deviations of the nominal environment parameters.

\subsection{Robustness to noise}
We also assessed the sensitivity of the model to the presence of sensor and phase noise. The localization model and the rotor phases were trained on a nominal environment (noiseless training) as well as in the presence of simulated noise injected in the relevant parameters (noisy training). 

\emph{Sensor noise} was emulated by adding white Gaussian noise of different amplitudes to the input sound. Signal-to-noise ratios (SNRs) ranging from $5$ dB to $\infty$ were evaluated. For noisy training, noise was injected in the range of $25-35$ dB SNR.

\emph{Phase noise} accounts for inexact control of the rotor phases that are not controlled exactly. We injected colored noise with SNRs ranging from $5$ dB to  $\infty$ simulating the effect of a PD controller. For noisy training in the presence of phase noise, the noise was only injected in the forward pass while being masked during backpropagation. Noisy training was performed at $15$ and $24$ dB SNR.

Localization accuracy in response to these perturbations in depicted in Fig. \ref{fig:robustness} (E-F). The model appears resilient to realistic levels of sensor and phase noise. As expected, noisy training improves robustness at the expense of mildly degraded performance in the noiseless setting.

\section{Discussion}\label{sec:discussion}

In this work, we introduced, to the best of our knowledge for the first time, a localization algorithm for multi-rotor aircraft relying on the propulsion noise produced by the drone's rotors. We demonstrate in simulation that the active shaping of the rotors phases substantially improves the localization accuracy and evaluate the algorithm robustness against various types of noise and modeling errors. We also provide a unique dataset of real rotor pressure field recordings in free space as well as a fully-differentiable forward model.
\newline

\emph{Limitations and future work.}
While conceptually extensible to three dimensions, all our simulations focused on the two-dimensional localization problem. The sensitivity of a predominantly flat pressure field to the vertical location will be assessed in future studies.
Our focus in this work was limited to localization within a known environment (up to some modeling uncertainties). The ability of the proposed approach to perform simultaneous localization and mapping (SLAM) is an exciting possibility left for future research.
Finally, except for the phase noise experiment, we assumed that the nominal phases are realized perfectly by the aircraft. In reality, the flight control system is required to trade-off between vehicle stability and the accuracy of the phase. The integration of the localization algorithms with a realistic phase controller is deferred to future research. 

\section*{ACKNOWLEDGMENT}
This project has received funding from the European Research Council (ERC) under the European Union’s Horizon 2020 research and innovation programme (grant agreement No. 863839). We are grateful to Yair Atzmon, Matan Jacoby, Aram Movsisian, and Alon Gil-Ad for their help with the data acquisition.

\addtolength{\textheight}{-12cm}   


\bibliographystyle{IEEEtran}
\bibliography{bibtex,IEEEabrv}

\end{document}